\documentclass[conference]{IEEEtran}
\IEEEoverridecommandlockouts
\usepackage{cite}
\usepackage{times}
\usepackage{soul}
\usepackage{url}
\usepackage[hidelinks]{hyperref}
\usepackage[utf8]{inputenc}
\usepackage[small]{caption}
\usepackage{graphicx}
\usepackage{amsmath}
\usepackage{booktabs}
\usepackage[english]{babel}
\usepackage[utf8]{inputenc}
\usepackage[autostyle]{csquotes}
\usepackage{mathtools}
\usepackage{amssymb}
\usepackage{bm}
\usepackage[T1]{fontenc}    % use 8-bit T1 fonts
\usepackage{lmodern}
\usepackage{hyperref}       % hyperlinks
\usepackage{url}            % simple URL typesetting
\usepackage{amsfonts}       % blackboard math symbols
\usepackage{multirow}
\usepackage{nicefrac}       % compact symbols for 1/2, etc.
\usepackage{microtype}      % microtypography
\usepackage{cleveref}
\graphicspath{{figures/}}
\usepackage[dvipsnames]{xcolor}
\usepackage{pgfplots}
\usepgfplotslibrary{groupplots}
\usetikzlibrary{matrix}
\usepackage{algorithm}
\usepackage{algorithmic}
\DeclareMathOperator{\argmin}{arg\,min}
\usepackage[colorinlistoftodos]{todonotes}

\DeclareMathOperator*{\argmax}{arg\,max}

\usepackage[caption=false,font=normalsize,labelfont=sf,textfont=sf]{subfig}
\definecolor{marine}{RGB}{0,32,96}
\definecolor{maroon}{RGB}{128,0,0}
\definecolor{olivegreen}{RGB}{85,107,47}
\definecolor{orange}{RGB}{229,148,0}

\pgfplotsset{every tick label/.append style={font=\tiny}}
\pgfplotsset{
	box plot width/.initial=4em,
	box plot/.style={
		/pgfplots/.cd,
		black,
		only marks,
		mark=-,
		mark size=\pgfkeysvalueof{/pgfplots/box plot width},
		/pgfplots/error bars/.cd,
		y dir=plus,
		y explicit,
	},
	box plot box/.style={
		/pgfplots/error bars/draw error bar/.code 2 args={%
			\draw [line width=0.20mm]  ##1 -- ++(\pgfkeysvalueof{/pgfplots/box plot width},0pt) |- ##2 -- ++(-\pgfkeysvalueof{/pgfplots/box plot width},0pt) |- ##1 -- cycle;
		},
		/pgfplots/table/.cd,
		y index=2,
		y error expr={\thisrowno{3}-\thisrowno{2}},
		/pgfplots/box plot
	},
	box plot top whisker/.style={
		/pgfplots/error bars/draw error bar/.code 2 args={%
			\pgfkeysgetvalue{/pgfplots/error bars/error mark}%
			{\pgfplotserrorbarsmark}%
			\pgfkeysgetvalue{/pgfplots/error bars/error mark options}%
			{\pgfplotserrorbarsmarkopts}%
			\path ##1 -- ##2;
		},
		/pgfplots/table/.cd,
		y index=4,
		y error expr={\thisrowno{2}-\thisrowno{4}},
		/pgfplots/box plot
	},
	box plot bottom whisker/.style={
		/pgfplots/error bars/draw error bar/.code 2 args={%
			\pgfkeysgetvalue{/pgfplots/error bars/error mark}%
			{\pgfplotserrorbarsmark}%
			\pgfkeysgetvalue{/pgfplots/error bars/error mark options}%
			{\pgfplotserrorbarsmarkopts}%
			\path ##1 -- ##2;
		},
		/pgfplots/table/.cd,
		y index=5,
		y error expr={\thisrowno{3}-\thisrowno{5}},
		/pgfplots/box plot
	},
	box plot median/.style={
		/pgfplots/box plot
	}
}
\pgfplotsset{yticklabel style={text width=1.5em, align=right}}
\pgfplotsset{compat=1.12}

\begin{document}

\title{Learning Primal Heuristics for Mixed Integer Programs}

\author{\IEEEauthorblockN{Yunzhuang Shen}
\IEEEauthorblockA{\textit{School of} \\ 
\textit{Computing Technologies}\\
\textit{RMIT University}\\
Melbourne, Australia \\
s3640365@student.rmit.edu.au}
\and
\IEEEauthorblockN{Yuan Sun}
\IEEEauthorblockA{\textit{School of Mathematics} \\
\textit{Monash University}\\
Melbourne, Australia \\
yuan.sun@monash.edu}
\and
\IEEEauthorblockN{Andrew Eberhard}
\IEEEauthorblockA{\textit{School of Science} \\
\textit{RMIT University}\\
Melbourne, Australia \\
andy.eberhard@rmit.edu.au}
\and
\IEEEauthorblockN{Xiaodong Li}
\IEEEauthorblockA{\textit{School of} \\ 
\textit{Computing Technologies}\\
\textit{RMIT University}\\
Melbourne, Australia \\
xiaodong.li@rmit.edu.au}
}

\maketitle

\begin{abstract}

% This paper proposes a novel primal heuristic for solving mixed integer programs (MIPs) by employing machine learning techniques.
This paper proposes a novel primal heuristic for Mixed Integer Programs, by employing machine learning techniques. Mixed Integer Programming is a general technique for formulating combinatorial optimization problems. Inside a solver, primal heuristics play a critical role in finding good feasible solutions that enable one to tighten the duality gap from the outset of the Branch-and-Bound algorithm (B\&B), greatly improving its performance by pruning the B\&B tree aggressively. In this paper, we investigate whether effective primal heuristics can be automatically learned via machine learning. We propose a new method to represent an optimization problem as a graph, and train a Graph Convolutional Network on solved problem instances with known optimal solutions. This in turn can predict the values of decision variables in the optimal solution for an unseen problem instance of a similar type. The prediction of variable solutions is then leveraged by a novel configuration of the B\&B method, Probabilistic Branching with guided Depth-first Search (PB-DFS) approach, aiming to find (near-)optimal solutions quickly. The experimental results show that this new heuristic can find better primal solutions at a much earlier stage of the solving process, compared to other state-of-the-art primal heuristics. 

\begin{IEEEkeywords}
Mixed Integer Programming, Primal Heuristics, Machine Learning
\end{IEEEkeywords}

\end{abstract}
\section{Introduction}
Combinatorial Optimization problems can be formulated as Mixed Integer Programs (MIPs). To solve general MIPs, sophisticated software uses Branch-and-Bound (B\&B) framework, which recursively decomposes a problem and enumerates over the sub-problems. A bounding function determines whether a sub-problem can be safely discarded by comparing the objective value of the current best solution (incumbent) to a dual bound, generally obtained by solving a linear programming relaxation (i.e., relaxing integral constraints) of that sub-problem. 
 
Inside solvers, primal heuristics play an important role in finding good primal solutions at an early stage \cite{achterberg2012rounding}. A good primal solution strengthens the bounding functions which allows one to prune suboptimal branches more aggressively \cite{berthold2006primal}. Moreover, finding good feasible solutions earlier greatly reduces the duality gap, which is important for user satisfaction \cite{fischetti2010heuristics}. Acknowledging the importance of primal heuristics, a modern open-source MIP solver SCIP \cite{achterberg2009scip}, employs dozens of heuristics \cite{berthold2006primal}, including meta-heuristics \cite{aarts2003local}, heuristics supported by mathematical theory \cite{berthold2014rens}, and heuristics mined by experts and verified by extensive experimental evidence \cite{achterberg2012rounding}. Those heuristics are triggered to run with engineered timings during the B\&B process.

In many situations, users are required to solve MIPs of a similar structure on a regular basis \cite{gasse2019exact}, so it is natural to seek Machine Learning (ML) solutions. In particular, a number of studies leverage ML techniques to speed up finding good primal solutions for MIP solvers. He et al. \cite{he2014learning} train Support Vector Machine (SVM) to decide whether to explore or discard a certain sub-problem, aiming to devote more of the computational budget on ones that are likely to contain an optimal solution. Khalil et al. \cite{khalil2017learning} train an SVM model to select which primal heuristic to run at a certain sub-problem. More related to our work here, Sun et al. \cite{sun2019using} and Ding et al. \cite{ding2019optimal} leverage ML to predict values of decision variables in the optimal solution, which are then used to fix a proportion of decision variables to reduce the size of the original problem, in the hope that the reduced space still contains the optimal solution of the original problem.

% with the hope of not pruning the optimal solution of the original problem. 
% which are then used to fix a proportion of decision variables and thus reduce the size of the original problem with the hope of not pruning the optimal solution of the original problem.
% In many situations, users are required to solve MIPs of a similar structure on a regular basis \cite{gasse2019exact}, so it is natural to seek machine learning solutions. However, it is not immediately apparent how to use machine learning to develop a general heuristic for MIP. We highlight some of the most important issues: first, although many heuristics can be adapted to a specific problem, it should be also applicable to general MIPs. Thus, one cannot assume problem-specific structural information or domain knowledge. Secondly, in the context of optimisation, the efficiency of a heuristic is also of crucial importance. A model with a high prediction time is not desirable \cite{gasse2019exact}. Thirdly, from a practical perspective, showing evidence that a machine learning model can generalize to unseen (larger) instances is also crucial. 

In this work, we propose a novel B\&B algorithm guided by ML, aiming to search for high-quality primal solutions efficiently. Our approach works in two steps. Firstly, we train an ML model using a dataset formed by optimally-solved small-scale problem instances where the decision variables are labeled by the optimal solution values. Specifically, we employ the Graph Convolutional Network \cite{kipf2016semi} (GCN), where an input graph associated with a problem instance to GCN is formed by representing each decision variable as a node and assigning an edge between two nodes if the corresponding decision variables appear in a constraint in the MIP formulation (see Section \ref{sec:mip}). Then given an unseen problem instance, the trained GCN with the proposed graph representation method can efficiently predict for each decision variable its probability of belonging to an optimal solution in an unseen problem instance (e.g., whether a vertex is a part of the optimal solution for the Maximum Independent Set Problem). The predicted probabilities are then used to guide a novel B\&B configuration, called Probabilistic Branching technique with guided Depth-first Search (PB-DFS). PB-DFS enumerates over the search space starting from the region more likely to contain good primal solutions to the region of unpromising ones, indicated by GCN.

Although both the problem-reduction approaches \cite{sun2019using,ding2019optimal} and our proposed PB-DFS utilize solution prediction by ML, they are inherently different. The former can be viewed as a pre-processing step to prune the search space of the original problem, and the reduced problem is then solved by a B\&B algorithm. In contrast, our PB-DFS algorithm configures the search order of the B\&B method itself and directly operates on the original problem. In this sense, our PB-DFS is an exact method if given sufficient running time. However, as the PB-DFS algorithm does not take into account the size of the B\&B search tree, it is not good at proving optimality. Therefore, we will limit the running time of PB-DFS and use it as a primal heuristic.

Our contributions can be summarized as follows:
\begin{enumerate}
    
    \item We propose the Probabilistic Branching technique with guided Depth-first Search, a configuration of the B\&B method specialized for boosting primal solutions empowered by ML techniques. 
    \item We propose a novel graph representation method that captures the relation between decision variables in a problem instance. The constructed graph can be input to GCN to make predictions efficiently. 
    \item Extensive experimental evaluation on NP-hard covering problems shows that 1) GCN with the proposed graph representation method is very competitive in terms of efficiency and effectiveness as compared to a tree-based model, a linear model, and a variant of Graph Neural Network \cite{ding2019optimal}. 2) PB-DFS can find (near-)optimal solutions at a much earlier stage comparing to existing primal heuristics as well as the problem-reduction approaches using ML.
    
    % GCN with the proposed graph representation is very competitive as compared to linear and tree-based models and is much more efficient than a variant of Graph Neural Network recently proposed by Ding et al. \cite{ding2019optimal}. 
    % We found our adapted version of Graph Convolutional Network \cite{kipf2016semi} (GCN) is much more effective than linear and tree-based models and is much more efficient than a variant of Graph Neural Network recently proposed by Ding et al. \cite{ding2019optimal}
    
\end{enumerate}

\section{Background}
\label{sec:background}
\subsection{MIP and MIP solvers}
\label{sec:mip}
MIP takes the form $ \argmin \{\bm{c}^{T}\bm{x}\,|\, \bm{x} \in \mathcal{F}\}$. For an MIP instance with $n$ decision variables, $\bm{c} \in \mathbb{R}^{n}$ is the objective coefficient vector. $\bm{x}$ denotes a vector of decision variables. We consider problems where the decision variables $\bm{x}$ are binary in this study, although our method can be easily extended to discrete domain \cite{nair2020solving}. $\mathcal{F}$ is the set of feasible solutions (search space), which is typically defined by integrality constraints and linear constraints $\bm{A}\bm{x} \leq \bm{b}$, where $\bm{A} \in \mathbb{R}^{m \times n}$ and $\bm{b} \in \mathbb{R}^{m}$ are the constraint matrix and constraint right-hand-side vector, respectively. $m$ is the number of constraints. The goal is to find an optimal solution in $\mathcal{F}$ that minimizes a linear objective function. 
 
\begin{algorithm}[t!]
 \caption{Branch-and-Bound Algorithm}
 \label{alg:bnb}
 \begin{algorithmic}[1]
    \REQUIRE a problem instance: $\mathcal{I}$;
    \STATE the node queue:  $\mathcal{L} \gets \{\mathcal{I}\}$; 
    \STATE the incumbent and its objective value: $\hat{x} \gets \varnothing$, $\hat{c} \gets \infty$;

    \WHILE{$\mathcal{L}$ is not empty}
       \STATE choose $Q$ from $\mathcal{L}$; $\mathcal{L} \gets \mathcal{L} \setminus Q$;
       \STATE solve the linear relaxation $Q^{LP}$ of Q;
       \IF{$Q^{LP}$ is infeasible}       
          \STATE go to Line 3 ;   
       \ENDIF
       \STATE denote the LP solution $\hat{x}^{LP}$ ; 
       \STATE denote the LP objective $\hat{c}^{LP}$ ;  
       \IF{$\hat{c}^{LP} \le \hat{c}$}                
            \IF{$\hat{x}^{LP}$ is feasible in Q} 
                \STATE $\hat{x} \gets \hat{x}^{LP}$; $\hat{c} \gets \hat{c}^{LP}$;
            \ELSE
                \STATE split $Q$ into subproblems $Q = Q_1 \cap ... \cap Q_n$ ;
                \STATE $\mathcal{L} \gets \mathcal{L} \cap \{Q_1 .. Q_n\}$;
            \ENDIF       
       \ENDIF
    \ENDWHILE
    \RETURN $\hat{x}$
\end{algorithmic}
\end{algorithm}

For solving MIPs, exact solvers employ B\&B framework as their backbone, as outlined in Algorithm \ref{alg:bnb}. There are two essential components in the B\&B framework, \textit{branching policy} and \textit{node selection strategy}. A \textit{node selection strategy} determines the next (sub-)problem (node) $Q$ to solve from the queue $\mathcal{L}$,  which maintains a list of all unexplored nodes. B\&B obtains a lower bound on the objective values of Q by solving its Linear Programming (LP) relaxation $Q^{LP}$. If the LP solution (lower bound) is larger than the objective $\hat{c}\equiv c^T \hat{x}$ of the incumbent $\hat{x}$ (upper bound), then the sub-tree rooted at node $Q$ can be pruned safely. Otherwise, this sub-tree possibly contains better solutions and should be explored further. If the LP solution $\hat{x}^{LP}$ is feasible in $Q$ and of better objective value, the incumbent is updated by $\hat{x}^{LP}$. Otherwise, $Q$ is decomposed into smaller problems by fixing a candidate variable to a possible integral value. The resulting sub-problems are added to the node queue. \textit{Branching policy} is an algorithm for choosing the branching variable from a set of candidate variables, which contains decision variables taking on fractional values in the solution $\hat{x}^{LP}$. Modern solvers implement sophisticated algorithms for both components, aiming at finding good primal solutions quickly while maintaining a relatively small tree size. See \cite{achterberg2009scip} for a detailed review.

% For a detailed review, we refer readers to \cite{achterberg2009scip}.

\subsection{Heuristics in MIP solvers}
\label{subsec:heuristic}
During computation, primal heuristics can be executed at any node, devoted to improve the incumbent, such that more sub-optimal nodes can be found earlier and thus pruned without further exploration. Berthold \cite{berthold2006primal} classifies these primal heuristics into two categories: start heuristics and improvement heuristics. Start heuristics aim to find feasible solutions at an early solving stage, while improvement heuristics build upon a feasible solution (typically the incumbent) and seek better solutions. All heuristics run on external memory (e.g., a copy of a sub-problem) and do not modify the structure of the B\&B tree. For a more comprehensive description of primal heuristics, we refer readers to \cite{berthold2006primal,fischetti2010heuristics}.
\section{Method}
\label{sec:method}

In Section \ref{sec:sp}, we describe how to train a machine learning model to predict the probability for each binary decision variable of its value in the optimal solution. In Section \ref{sec:pb-dfs},  we illustrate the proposed PB-DFS that leverages the predicted probabilities to boost the search for high-quality primal solutions. 

\subsection{Solution Prediction}
\label{sec:sp}
Given a combinatorial optimization problem, we construct a training dataset from multiple optimally-solved problem instances where each instance associates with an optimal solution. A training example corresponds to one decision variable $x_i$ from a solved problem instance. The label $y_i$ of $x_i$ is the solution value of $x_i$ in the optimal solution associated with a particular problem instance. The features $\bm{f}_i$ of $x_i$ are extracted from the MIP formulation, which describes the role of $x_i$ in that problem instance. We describe those features in Appendix A. Given the training data, an ML model can be trained by minimising the cross-entropy loss function \cite{bishop2006pattern} to separate the training examples with different class labels \cite{sun2019using,ding2019optimal}. 

\begin{algorithm}[t!]
\caption{An MIP Instance to Linkage Graph}
 \label{alg:gg}
 
 \begin{algorithmic}[1]
 \REQUIRE the constraint matrix: $A \in \mathbb{R}^{m \times n}$;
 \STATE the adjacency matrix: $\mathcal{G}^{adj} \gets \boldsymbol{0}^{n \times n}$;
 \STATE the row index of $A$: $i \gets 0$;
 \STATE the index set of variables: $\mathcal{C} \gets \varnothing$;
 \WHILE{$i < m$}
 \STATE $\mathcal{C} \gets \{j \;|\; A_{i,j} \neq 0\} $;
      \FOR{$k,l \in \mathcal{C} \:,\: k \neq l$}
            \STATE   $\mathcal{G}^{adj}_{k,l} \gets 1$, $\mathcal{G}^{adj}_{l,k} \gets 1$;
      \ENDFOR
      \STATE $ i \gets i + 1$;
  \ENDWHILE
     \RETURN{$\mathcal{G}^{adj}$}
 \end{algorithmic}
\end{algorithm}

% \begin{figure}[t!]
%  \centering
%  \resizebox{0.9\columnwidth}{!}{
% \begin{algorithm2e}
% \KwIn{the constraint matrix: $A \in \mathbb{R}^{m \times n}$ }
% \textbf{initialization}: the adjacency matrix: $\mathcal{G}^{adj} \gets \boldsymbol{0}^{n \times n}$\; \hspace{6.7em} the row index of $A$: $i \gets 0$\; \hspace{6.7em} the index set of variables $\mathcal{C} \gets \varnothing $ \;
%     \While{$i < m$}{
%       $\mathcal{C} \gets \{j \;|\; A_{ij} \neq 0\} $\;
%       \For{$k,l \in \mathcal{C} \:,\: k \neq l$}{
%               $\mathcal{G}^{adj}_{k,l} \gets 1$, $\mathcal{G}^{adj}_{l,k} \gets 1$
%       }
%       $ i \gets i + 1$
%     }
    
%     \Return{$\mathcal{G}^{adj}$}
% \end{algorithm2e}
% }
% \end{figure}

We adapt the Graph Convolutional Network (GCN) \cite{kipf2016semi} for this classification task, a type of Graph-based Neural Network (GNN), to take the relation between decision variables from a particular problem instance into account. To model the relation between decision variables, we propose a simple method to extract information from the constraint matrix of an MIP. Algorithm \ref{alg:gg} outlines the procedure. Given an MIP we represent each decision variable as a node in a graph, and assign an edge between two nodes if the corresponding decision variables appear in a constraint. This graph representation method can capture the linkage of decision variables effectively, especially for graph-based problems. For example, the constructed graph for the Maximum Independent Set problem is exactly the same as the graph on which the problem is defined, and the constructed graph for the Traveling Salesman Problem is the line graph of the graph given by the problem definition.

%Intuitively, this graph representation method can well capture the characteristics of graph-based problems, e.g., the constructed graph for Maximum Independent Set problem is a generic graph, and for Traveling Salesman Problem, the generated graph is the line graph \cite{harary1960some} of a generic graph.\comment[id=YS]{I don't think ``generic graph'' makes much sense. Perphas replace it as ``This graph representation method can well capture the characteristics of MIPs especially graph-based problems. For example, the constructed graph for the Maximum Independent Set problem is the same as the graph on which the problem is defined, and the constructed graph for the Traveling Salesman Problem is the line graph of the graph given by the problem definition.''} 

Given the dataset containing training examples $(\bm{f}_i, y_i)$ grouped by problem instances and each problem instance associated with an adjacency matrix $\mathcal{G}^{adj}$ representing the relation between decision variables, we can then train GCN. Specifically for a problem instance, the adjacency matrix $\mathcal{G}^{adj}$ is precomputed to normalize graph Laplacian by 
\begin{equation}
  L \coloneqq I - D^{-\frac{1}{2}}\mathcal{G}^{adj}D^{-\frac{1}{2}},  
\end{equation}
where $I$ and $D$ are the identity matrix and diagonal matrix of $\mathcal{G}^{adj}$, respectively. The propagation rule is defined as 
\begin{equation}
  H^{l+1} \coloneqq \sigma(LH^{l}W^{l} + H^{l}),  
\end{equation}
where $l$ denotes the index of a layer. Inside the activation function $\sigma(\cdot)$,  $W^{l}$ denotes the weight matrix. $H^l$ is a matrix that contains hidden feature representations for decision variables in that problem instance, initialized by the feature vectors of decision variables $H^0 = [\bm{f}_1, \cdots, \bm{f}_{n}]^{T}$. The second term is referred to as the residual connection, which preserves information from the previous layer. For hidden layers with arbitrary number of neurons, We adopt $ReLU(x) = \max(x,0)$ as the activation function \cite{nair2010rectified}. For the output layer $L$, there is only one neural to output a scalar value for each decision variable and sigmoid function is used as the activation function for prediction $H_{L}= [\hat{y}_1, \cdots, \hat{y}_n]$. We train GCN using Stochastic Gradient Descent to minimize the cross-entropy loss function between the predicted values $\hat{y}_i$ of decision variables and their optimal values $y$, defined as 
\begin{equation}
    \min \; -\frac{1}{N}\sum_{i=1}^{N}\big(y_i \times \log(\hat{y}_i) + (1-y_i) \times \log(1-\hat{y}_i)\big),
\end{equation}
where $N$ is the total number decision variables from multiple training problem instances.

Given an unseen problem instance at test time, we can use the trained GCN to predict for each decision variable $x_i$ a value $\hat{y}_i$, which can be interpreted as the probability of a decision variable taking the value of $1$ in the optimal solution $ p_i = P(x_i=1)$.  We refer to the array of predicted values as the \textit{probability vector}.

\subsection{Probabilistic Branching with guided Depth-first Search} 
\label{sec:pb-dfs}

%h, which can potentially be used as an effective primal heuristic to find a high-quality solution quickly

We can then apply the predicted \textit{probability vector} to guide the search process of B\&B method. The proposed \textit{branching strategy}, Probabilistic Branching (PB) attempts to greedily select variable $x_i$ from candidate variables with the highest score $z_i$ to branch on. The score of variable $x_i$ can be computed by 
\begin{equation}
    z_i \gets \max(p_i, 1-p_i),
\end{equation}
where $p_i$ is the probability of $x_i$ being assigned to $1$ predicted by the GCN model. This function assigns a higher score to variables whose $p_i$ is closer to either 0 or 1. This value tells how certain an ML model is about its prediction. We then can branch on the decision variable $x_i$ with the highest score,
\begin{gather}
    i \gets \argmax_i z_i; \;\;s.t.\; i \in \mathcal{C}.    
\end{gather}
$\mathcal{C}$ is the index set of candidate variables that are not fixed at the current node. In this way, our PB method prefers to branch on the variables for which our prediction is more ``confident'' at the shallow level of the search tree while exploring the uncertain variables at the deep level of the search tree.

We propose to use a guided Depth-first Search (DFS) as the \textit{node selection strategy} to select the next sub-problem to explore. After branching a node, we have a maximum of two child nodes to explore (because the decision variables are binary). Guided DFS selects the child node that results from fixing the decision variable to the nearest integer of $p_i$. When reaching a leaf node, guided DFS backtracks to the deepest node in the search tree among all unexplored nodes. Therefore, guided DFS always explores the node most likely to contain optimal solutions, instructed by the prediction of an ML model. We refer to this implementation of B\&B method as PB-DFS.

\begin{figure}[t!]
\centering
\includegraphics[width=0.8\linewidth]{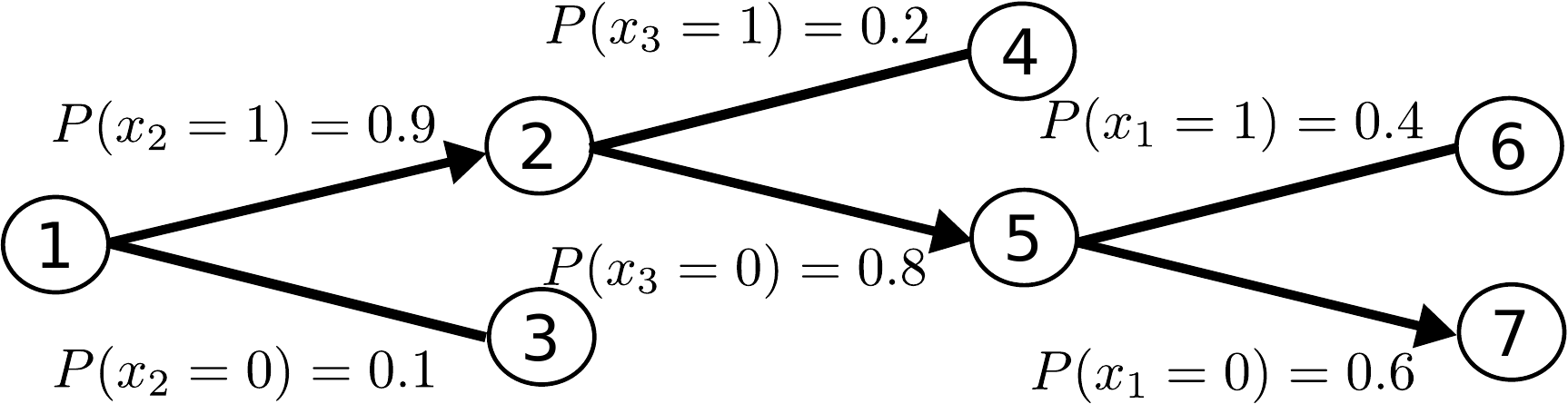}
\caption{Probabilistic Branching with guided Depth-first Search on three decision variables. Given a node, PB branches on a variable our prediction is more confident. The variables that have already been branched are removed from the candidate set. The order of node selection by guided DFS is indicated by the arrow lines.}
\label{fig:tree}
\end{figure}

Figure \ref{fig:tree} illustrates PB-DFS applied to a problem with three decision variables. We note that as a configuration of the B\&B framework, PB-DFS can be an exact search method if given enough time. However, it is specialized for aggressively seeking high-quality primal solutions while trading off the size of the B\&B tree created during the computation. Hence, we implement and evaluate it as a primal heuristic, which partially solves an external copy of the (sub-)problem with a certain termination criterion.

% Note that PB-DFS can be deployed in alternative ways. For instance, one can use PB-DFS at the start of the problem solving to boost primal solutions and then switch to other B\&B configurations that focus on improving the dual bound.

\section{Experiments}
\label{sec:exp}

In this section, we use numerical experiments to evaluate the efficacy of our proposed method. After describing the experiment setup, we first analyse different ML models in terms of both effectiveness and efficiency. Then, we evaluate the proposed PB-DFS equipped with different ML models against a set of primal heuristics. Further, we show the significance of PB-DFS with the proposed GCN model by comparing it to the full-fledged SCIP solver and problem-reduction approaches using the SCIP solver \cite{ding2019optimal}.

\subsection{Setup}
 
\paragraph{Test Problems} We select a set of representative NP-hard problems: Maximum Independent Set (MISP), Dominant Set Problem (DSP), Vertex Cover Problem (VCP), and an additional problem from Operational Research, Combinatorial Auction Problem (CAP).  For each problem, we generate instances of three different scales, i.e., small, medium, and large. Small-scale problem instances and medium-scale problem instances are solved to optimality for training and evaluating ML models. Large-scale problem instances are used for evaluating different solution approaches. Details of problem formulations and instance generations are provided in Appendix B. 

\paragraph{ML Models}  We refer to the GCN that takes the proposed graph representation of MIP as LG-GCN, where LG stands for Linkage Graph. We compare LG-GCN against three other machine learning (ML) models, Logistic Regression (LR), XGBoost \cite{chen2016XGBoost}, and a variant of Graph Neural Network (GNN) which represents the constraint matrix of MIP as a tripartite graph \cite{ding2019optimal}. We address this GNN variant as TRIG-GCN, where TRIG stands for Tripartite Graph. We set the number of layers to $20$ for LG-GCN and that of TRIG-GCN is set to $2$ due to its high computational cost (explained later). For these two Graph Neural Networks, the dimension of the hidden vector of a decision variable is set to $32$. For LR and XGBoost, the hyper-parameters are the default ones in the Scikit-learn \cite{pedregosa2011scikit} package. For all ML models, the feature vector of a decision variable has 57 dimensions, containing statistics extracted from the MIP formulation of a problem instance (listed in Appendix A). For each feature, we normalize its values to the range $[0, 1]$ using min-max normalization with respect to the decision variables in a particular problem instance. Besides, LG-GCN and TRIG-GCN are trained with graphs of different structures with respect to each problem instance. For a problem, an ML model is trained using $500$ optimally-solved small-scale instances.

\paragraph{Evaluation of ML Models}  We evaluate the classification performance of ML models on two test datasets, constructed from $50$ small problem instances (different from the training instances) and medium-sized problem instances, respectively. We measure a model's performance with the Average Precision (AP) \cite{zhu2004recall}, defined by accumulating the product of precision and the change in recall when moving the decision threshold on a set of ranked variables. AP is a more informative metric than the accuracy metric in our context, because it takes the ranking of decision variables into account. This allows AP to better measure the classification performance for imbalanced data that is common in the context of solution prediction for NP-hard problems. Further, AP can better measure the performance of the proposed PB-DFS, because exploring a node not containing the optimal solution in the upper level of the B\&B tree is more harmful than exploring one in the lower level of the tree.

% We measure a model's performance with the Average Precision (AP) \cite{zhu2004recall}, defined by $\sum_{k=1}^n P(k)\Delta r(k)$ over a number of $n$ ranked variables. $P(k)$ denotes the precision when taking the score of $k^{th}$ variable as a decision threshold. $\Delta r(k)$ stands for the difference between recalls at the ${k}^{th}$ and ${(k-1)}^{th}$ thresholds.

\begin{table*}[t!]
\centering
\caption{Comparison between ML models. AP column shows the mean statistic of Average Precision values over $50$ problem instances. We conduct student's $t$-test by comparing LG-GCN against other baselines. $p$-value less than $0.05$ (a typical significance level) indicates that the LG-GCN is significantly better than other ML models.}
\label{tab:ml}
\resizebox{0.95\textwidth}{!}{

\begin{tabular}{@{}cccccccccc@{}}
\toprule
	  \multirow{2}{*}{Problem Size} & \multirow{2}{*}{Model}	  &   \multicolumn{2}{c}{Independent Set} & \multicolumn{2}{c}{Dominant Set} &   \multicolumn{2}{c}{Vertex Cover} & \multicolumn{2}{c}{Combinatorial Auction}\\
\cline{3-10}

& & AP & $p$-value & AP & $p$-value &  AP & $p$-value & AP & $p$-value \\
\midrule

\multirow{4}{*}{Small} &LG-GCN  & \textbf{96.53} & - & 87.25 & - & \textbf{98.05} & - & \textbf{46.10} & - \\
					   &TRIG-GCN & 88.62 & 8.7E-28  & 86.90 & 8.0E-02 & 92.42 & 2.8E-87 &41.77 & 4.7E-02  \\
						&XGBoost & 74.11 & 3.9E-135 & 86.45 & 2.6E-01 & 84.82 & 1.7E-140 & 39.05 & 1.4E-03 \\
					    &LR & 74.24 & 1.6E-128 & 86.59 & 3.5E-01 & 85.29 & 2.2E-137 & 41.31 &2.3E-02 \\

\midrule
                        
\multirow{4}{*}{Medium} &LG-GCN & \textbf{96.41} & - & 87.52 & - & \textbf{98.24} & - & \textbf{47.07} & - \\
						&TRIG-GCN & 88.16 & 4.8E-31 & 87.18 & 5.1E-02 & 92.74 &2.1E-51 & 42.67 & 6.0E-04 \\
						&XGBoost & 73.07 & 1.1E-125 & 86.80 & 2.0E-01 & 84.70 & 5.2E-67 & 40.85 & 1.2E-06\\
					    &LR & 73.99 & 2.0E-148 & 86.86 & 2.3E-01 & 85.33 & 3.3E-57 & 42.35 & 1.8E-04 \\

\bottomrule
\end{tabular}}
\end{table*}

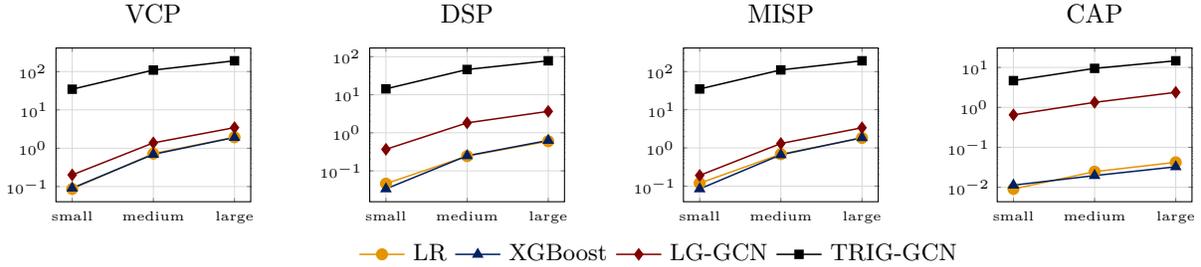
\begin{figure*}[t!]
    \centering
    \begin{tikzpicture}
    \begin{groupplot}[group style = {group size = 4 by 1, horizontal sep = 45pt, vertical sep = 40pt}, height=0.2\textwidth,width=0.23\textwidth, ymode=log, grid style={line width=.1pt, draw=gray!10},major grid style={line width=.2pt,draw=gray!30}, xmajorgrids=true, ymajorgrids=true,  major tick length=0.05cm, minor tick length=0.0cm, xtick=data,xticklabels={small, medium, large}, legend style={column sep = 1pt, legend columns = 4,font=\small,draw=none}]
	 \nextgroupplot[title=  VCP , ytick={0.1,1,10,100,1000}]
    \addplot[color=orange, mark=*, mark size = 2.0, line width=0.20mm] coordinates {
    (1,0.08679)(2,0.72263)(3,1.93152)
    };
    \addplot[color=marine, mark=triangle*, mark size=2pt, line width=0.20mm] coordinates {
    (1,0.09095)(2,0.691780)(3,1.90768)
    };
    \addplot[color=maroon,  mark= diamond*, mark size=2pt, line width=0.20mm] coordinates {
    (1,0.20097)(2,1.37839)(3,3.44374)
    };
    \addplot[color=black,  mark= square*, mark size=1.5pt, line width=0.20mm] coordinates {
    (1,34.61548)(2,109.20377)(3,190.78077)
    }; 
	 \nextgroupplot[title=  DSP, ytick={0.1,1,10,100,1000}]
    \addplot[color=orange, mark=*, mark size = 2.0, line width=0.20mm] coordinates {
    (1,0.04602)(2,0.24363)(3,0.60040)
    };
    \addplot[color=marine, mark=triangle*, mark size=2pt, line width=0.20mm] coordinates {
    (1,0.03352)(2,0.24906)(3,0.63001)
    };
    \addplot[color=maroon,  mark= diamond*, mark size=2pt, line width=0.20mm] coordinates {
    (1,0.36788)(2,1.82325)(3,3.65459)
    };
    \addplot[color=black,  mark= square*, mark size=1.5pt, line width=0.20mm] coordinates {
    (1,14.27948)(2,46.04907)(3,77.96740)};
    \nextgroupplot[title=   MISP, ytick={0.1,1,10,100,1000}]
    \addplot[color=orange, mark=*, mark size = 2.0, line width=0.20mm] coordinates {
    (1,0.11914)(2,0.68612)(3,1.82378)
    };
    \addplot[color=marine, mark=triangle*, mark size=2pt, line width=0.20mm] coordinates {
    (1,0.08431)(2,0.66030)(3,1.84929)
    };
    \addplot[color=maroon,  mark= diamond*, mark size=2pt, line width=0.20mm] coordinates {
    (1,0.19030)(2,1.30499)(3,3.37232)
    };
    \addplot[color=black,  mark= square*, mark size=1.5pt, line width=0.20mm] coordinates {
    (1,34.97686)(2,110.43061)(3,191.52347)}; 
    \nextgroupplot[title=  CAP, ytick={0.01, 0.1,1,10,100},    legend to name=grouplegend]
    \addplot[color=orange, mark=*, mark size = 2.0, line width=0.20mm] coordinates {
    (1,0.00909)(2,0.02458)(3,0.04208)
    };\addlegendentry{LR}
    \addplot[color=marine, mark=triangle*, mark size=2pt, line width=0.20mm]coordinates {
    (1,0.01128)(2,0.01959)(3,0.03251)
    };\addlegendentry{XGBoost}
    \addplot[color=maroon,  mark= diamond*, mark size=2pt, line width=0.20mm] coordinates {
    (1,0.64930)(2,1.33676)(3,2.37771)
    }; \addlegendentry{LG-GCN}
    \addplot[color=black,  mark= square*, mark size=1.5pt, line width=0.20mm] coordinates {
    (1,4.66971)(2,9.47864)(3,14.66900)
    };\addlegendentry{TRIG-GCN}
    \end{groupplot} 
    \node at (group c1r1.south) [anchor=west, yshift=-0.7cm, xshift=2.5cm] {\ref{grouplegend}};       
    \end{tikzpicture}
\caption{ Increase of model prediction time when enlarging the size of problem instances: the $y$-axis is the prediction time in seconds in log-scale, and the $x$-axis is the size of problem instances in $3$ scales.}
\label{fig:time}
\end{figure*}

\paragraph{Evaluation of Solution Methods}  We evaluate PB-DFS equipped with the best ML model against heuristic methods as well as problem-reduction approaches on large-scale problem instances. In the first part, PB-DFS is compared with primal heuristics that do not require a feasible solution on large problem instances. By examining the performance of the heuristics enabled by default in SCIP, four types of heuristics are selected as baselines: the Feasibility Pump \cite{fischetti2005feasibility} (FP), the Relaxation Enhanced Neighborhood Search \cite{berthold2006primal} (RENS), a set of 15 diving heuristics, and a set of 8 rounding heuristics \cite{achterberg2012rounding}. We allow PB-DFS to run only at the root node and terminate it upon the first-feasible solution is found. The compared heuristic methods run multiple times during the B\&B process under a cutoff time of 50 seconds with default running frequency and termination criteria tuned by SCIP developers. Generating cutting planes is disabled to best measure the time of finding solutions by different heuristics. In the second part, we demonstrate the effectiveness of PB-DFS by comparing SCIP solver with only PB-DFS as the primal heuristic against full-fledged SCIP solver where all heuristics are enabled as well as a problem-reduction approach by Ding et al. \cite{ding2019optimal}. The problem-reduction approach splits the root node of the search tree by a constraint generated from \textit{probability vector}, and then solved by SCIP-DEF. To alleviate the effects of ML predictions, we use the \textit{probability vector} generated by LG-GCN for both PB-DFS and ML-Split. The cutoff time is set to $1000$ seconds. Note that for DSP, we employ an alternative score function  $z_i \gets p_i$. The corresponding DFS selects the child node that is the result of fixing the decision variable to one when those nodes are at the same depth. A comparison of alternative score functions is detailed in Appendix C.

\paragraph{Experimental Environment} We conduct experiments on a cluster with 32 Intel 2.3 GHz CPUs and 128 GB RAM. PB-DFS is implemented using C-api provided by the state-of-the-art open-source solver, SCIP, version 6.0.1. The implementations of Logistic Regression and XGBoost are taken from Scikit-learn \cite{pedregosa2011scikit}. Both LG-GCN and TRIG-GCN are implemented using Tensorflow package \cite{abadi2016tensorflow} where offline training and online prediction are done by multiple CPUs in parallel. All solution approaches are evaluated on a single CPU core. Our code is available online\footnote{Code is available at \url{https://github.com/Joey-Shen/pb-dfs}.}.
% Generating cutting planes is disabled when evaluating the performance of these heuristics to best reveal their impact on the duality bound.

\subsection{Results on Solution Prediction}
\label{subsec:co}

Table \ref{tab:ml} presents the ML models' performance on solution prediction. The small-scale test instances are of the same size as the training instances and the medium-scale ones are used for examining the generalization performance of tested ML models. We cannot measure the classification performance of ML models on large-scale instances because the optimal solutions for them are not available. By comparing the mean statistic of AP values, we observe that LG-GCN is very competitive among all problems indicated by mean statistics. We conduct student's $t$-test by comparing LG-GCN against other baselines where the AP values for a group of test problem instances of LG-GCN is compared with that of other ML models. In practice, $p$-value less than $0.05$ (a typical significance level) indicates that the difference between the two samples is significant. Therefore, we confirm that the proposed LG-GCN can predict solutions with better quality as compared to other ML models on MISP, VCP, and CAP. The performances of ML models on DSP are comparable. Note that on CAP which is not originally formulated on graphs, LG-GCN's AP value is significantly better than TRIG-GCN's. This shows that the proposed graph construction method is more robust when extending to non-graph based problems as compared to TRIG-GCN. LR is slightly better than XGBoost overall. All models show their capability to generalize to larger problem instances.

In addition to prediction accuracy, the prediction time of an ML model is a part of the total solving time, which should also be considered for developing efficient solution methods. Figure \ref{fig:time} shows the increase of the prediction time when enlarging the problem size for the compared ML models. Comparing graph-based models, we observe that in practice the mean prediction time by LG-GCN is much less than the one by TRIG-GCN. The high computational cost of TRIG-GCN prevents us from building it with a large number of layers. The computation time of LG-GCN is close to those of linear models on VCP and MISP and shifts away on DSP and CAP. This is understandable because the complexity of LG-GCN is linear in the number of edges in a graph \cite{kipf2016semi} and is polynomial with respect to the number of decision variables, as compared to linear growth for LR and XGBoost. However, for MISP and VCP where the constraint coefficient matrix is sparse (i.e., the fraction of zero values is high), in practice, the difference in the growth of computation time with increasing problem size is not as dramatic, but it may be significant when an MIP instance has a dense constraint matrix, e.g., Combinatorial Auction Problems.

\subsection{Results For Finding Primal Solutions}

\begin{table}[t!]
\centering
\caption{Comparison between PB-DFS and primal heuristics.}
\label{tab:heuristics}
\resizebox{\columnwidth}{!}{
\begin{tabular}{ccccccc}
\toprule
 Problem & Heuristic & \begin{tabular}{@{}c@{}}Best Solution \\ Objective\end{tabular} &  \begin{tabular}{@{}c@{}}Best Solution \\ Time\end{tabular} & \begin{tabular}{@{}c@{}}\# Instances \\ no feasible solution\end{tabular} & \# Calls & \begin{tabular}{@{}c@{}}Heuristic \\ Total Time \end{tabular} \\
\midrule
 \multirow{4}{*}{VCP (Min.)} & FP & 2137.3 & 1.6  & 19 & 1.0 &1.0 \\
&Roundings      	& 1784.7 & 35.2 & 0 &  252.0 & 8.0 \\ 
&PB-DFS-LR			 & 1634.0 & 3.8	 & 0 & 1.0 & 3.8 \\
&PB-DFS-GCN			 & \textbf{1629.0} (\textbf{1628.2}) & 5.7 (6.0) & 0 & 1.0 & 5.7 (21.6)\\\cmidrule{1-7}

\multirow{5}{*}{DSP (Min.)} &FP & 325.1 & 4.0  & 2 & 1.0 & 0.4 \\
&Roundings      	& 515.1 & 47.9 & 0 &  135.5 & 0.8 \\ 
&RENS      	 & 320.6 & 19.6 & 17 & 1.0 & 13.6\\ 
&PB-DFS-LR			 & \textbf{318.4} & 9.0	 & 0 & 1.0 & 9.0 \\
&PB-DFS-GCN			 & \textbf{318.7} (\textbf{316.1}) & 10.6 (23.6) & 0 & 1.0 & 10.6 (21.6)\\\midrule

 \multirow{4}{*}{MISP (Max.)} & FP & 845.8 & 1.3  & 2 & 1.0 &0.9 \\
&Roundings      	& 1225.6 & 38.0 & 0 &  681.0 & 18.0 \\ 
&Divings      	 & 1260.0 & 41.5 & 0 & 9.0 & 6.6\\ 
&PB-DFS-LR			 & 1365.8 & 3.9	 & 0 & 1.0 & 3.9 \\
&PB-DFS-GCN			 & \textbf{1371.0} (\textbf{1371.6})& 4.5 (5.6) & 0 & 1.0 & 4.5 (21.4)\\\cmidrule{1-7}

\multirow{5}{*}{CAP (Max.)}& FP	& - & -  & 30 & 1.0 & 0.3 \\
&Divings      	 & \textbf{3633.2} & 26.1 & 0 & 36.2 & 6.1\\ 
&Roundings      	& 3274.4 & 20.4 & 0 & 990.5 & 0.25 \\
&RENS      	 & 3425.9 & 3.2 & 0 & 1.0 & 1.9\\ 
&PB-DFS-LR			 & 3018 & 4.1	 & 0 & 1.0 & 4.1 \\
&PB-DFS-GCN			 & 3280.5 (3582.8) & 4.5 (12.9) & 0 & 1.0 & 4.5 (22.3)\\

\bottomrule

\end{tabular}}
\end{table}

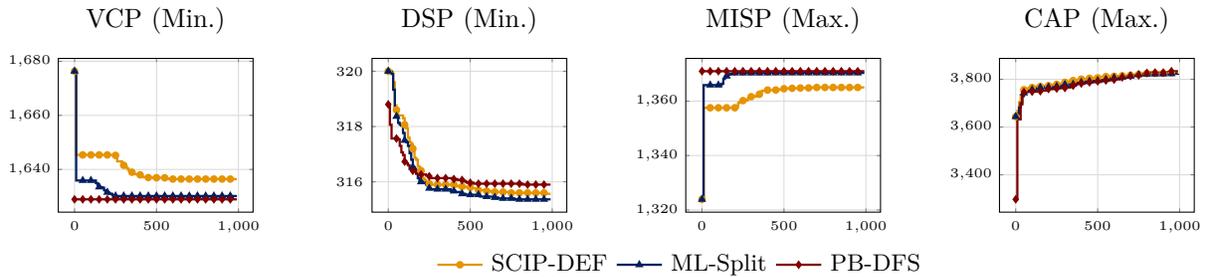
\begin{figure*}[t]
    \centering
	\begin{tikzpicture}
	\begin{groupplot}[group style = {group size = 4 by 1, horizontal sep = 45pt, vertical sep = 40pt}, height=0.2\textwidth,width=0.23\textwidth, grid style={line width=.1pt, draw=gray!10},major grid style={line width=.2pt,draw=gray!30}, xmajorgrids=true, ymajorgrids=true,  major tick length=0.05cm, minor tick length=0.0cm, legend style={column sep = 1pt, legend columns = 4,font=\small,draw=none}]
	 \nextgroupplot[title=  VCP (Min.),mark repeat=5,mark phase=0]
	\addplot[color=orange, mark=*, mark size = 1.0, line width=0.30mm, const plot] table [x=x, y=y, col sep=comma] {data/vc_scip_def.csv};
	\addplot[color=marine, mark=triangle*, mark size=1pt, const plot, line width=0.30mm] table [x=x, y=y, col sep=comma] {data/vc_ml_ding.csv};
	\addplot[color=maroon,  mark= diamond*, mark size=1pt, const plot, line width=0.30mm] table [x=x, y=y, col sep=comma] {data/vc_pb_dfs.csv};
	 \nextgroupplot[title=  DSP (Min.),mark repeat=5,mark phase=0]
	\addplot[color=orange, mark=*, mark size = 1.0, line width=0.30mm, const plot] table [x=x, y=y, col sep=comma] {data/ds_scip_def.csv};
	\addplot[color=marine, mark=triangle*, mark size=1pt, const plot, line width=0.30mm] table [x=x, y=y, col sep=comma] {data/ds_ml_ding.csv};
	\addplot[color=maroon,  mark= diamond*, mark size=1pt, const plot, line width=0.30mm] table [x=x, y=y, col sep=comma] {data/ds_pb_dfs.csv};
    \nextgroupplot[title=   MISP (Max.),mark repeat=5,mark phase=0]
	\addplot[color=orange, mark=*, mark size = 1.0, line width=0.30mm, const plot] table [x=x, y=y, col sep=comma] {data/mis_scip_def.csv};
	\addplot[color=marine, mark=triangle*, mark size=1pt, const plot, line width=0.30mm] table [x=x, y=y, col sep=comma] {data/mis_ml_ding.csv};
	\addplot[color=maroon,  mark= diamond*, mark size=1pt, const plot, line width=0.30mm] table [x=x, y=y, col sep=comma] {data/mis_pb_dfs.csv};
    \nextgroupplot[title=   CAP (Max.), mark repeat=5,mark phase=0, legend to name=grouplegend]
	\addplot[color=orange, mark=*, mark size = 1.0, line width=0.30mm, const plot] table [x=x, y=y, col sep=comma] {data/ca_scip_def.csv};\addlegendentry{SCIP-DEF}
	\addplot[color=marine, mark=triangle*, mark size=1pt, const plot, line width=0.30mm] table [x=x, y=y, col sep=comma] {data/ca_ml_ding.csv};\addlegendentry{ML-Split}
	\addplot[color=maroon,  mark= diamond*, mark size=1pt, const plot, line width=0.30mm] table [x=x, y=y, col sep=comma] {data/ca_pb_dfs.csv};\addlegendentry{PB-DFS}
    \end{groupplot} 
    \node at (group c1r1.south) [anchor=west, yshift=-0.7cm, xshift=3.5cm] {\ref{grouplegend}};       
    \end{tikzpicture}
\caption{Change of the primal bound during computation on large-scale problem instances: the $x$-axis is the solving time in seconds, and the $y$-axis is the objective value of the incumbent.}
\label{fig:primalbound}
\end{figure*}

Table \ref{tab:heuristics} shows the computational results of PB-DFS as compared to the most effective heuristic methods used in the SCIP solver. Recall that the FP and RENS are two standalone heuristics. \textit{Roundings} refers to a set of 8 rounding heuristics and \textit{Divings} covers 15 diving heuristics. PB-DFS-LR and PB-DFS-GCN stand for the PB-DFS method equipped with LG-GCN and LR model, respectively. Note that the PB-DFS method only runs at the root node and terminates upon finding the first feasible solution. Besides, we analyze an additional criterion using PB-DFS-GCN, terminating with a cutoff time $20$ seconds (statistics are shown in brackets). For each problem, we run a heuristic 30 times on different instances. Since the SCIP solver assigns heuristics with different running frequencies based on the characteristic of a problem, we only show heuristics called at least once per instance. The column \textit{\# Instances no feasible solution} reports the number of instances that a heuristic does not find a feasible solution. Other columns show statistics with a geometric mean shifted by one averaged over instances where a heuristic finds at least one feasible solution. Note that we only consider the best solutions found by a heuristic. Solutions found by branching are not included. We also show the average number of calls of a heuristic and the total running time by a heuristic in the last columns for reference. For those PB-DFS heuristics, mean prediction time by models is added to the \textit{Best Solution Time} and \textit{Heuristic Total Time}. Note that the primal heuristics do not meet the running criteria by SCIP for a certain problem is excluded from the results.

Overall, the PB-DFS methods can find better primal solutions much earlier on VCP, DSP, and MISP. It is less competitive on CAP. This is understandable because the AP value of the prediction for CAP is low (Table \ref{tab:ml}), indicating the prediction of an ML model is less effective. Comparing PB-DFS with different ML models, PB-DFS-GCN can find better solutions than PB-DFS-LR on VCP, MISP, and CAP, and they are comparable on DSP. This means that the AP values of \textit{probability vectors} for different models can reflect the performance of PB-DFS heuristics to a certain extent. When comparing two termination criteria using the PB-DFS-GCN model, we observe that giving the proposed heuristic more time can lead to better solutions. These results show that the PB-DFS method can be a very strong primal heuristic when the ML prediction is of high-quality (i.e., high AP values).

% Rounding heuristics are called frequently by SCIP in all problems, but the quality of the solutions found is not always satisfactory. Similarly, FP heuristic can find good feasible solutions on DSP, but does not work well on other problems. Although diving heuristics are the best heuristics on CAP, they are either inferior to PB-DFS heuristics or do not meet the running criteria by SCIP on other problems.

\begin{table}[t!]
\centering
\caption{PB-DFS compared with SCIP-DEF and ML-Split.}
\label{tab:heuristics2}
\resizebox{0.98\columnwidth}{!}{
\begin{tabular}{cccccc}
\toprule
Problem & Method & \begin{tabular}{@{}c@{}}Best Solution \\ Objective\end{tabular} & \begin{tabular}{@{}c@{}}Best Solution \\ Time\end{tabular} & \begin{tabular}{@{}c@{}}Optimality \\ Gap (\%)\end{tabular} & \begin{tabular}{@{}c@{}}Heuristic \\ Total Time \end{tabular}  \\
\midrule

\multirow{3}{*}{VCP (Min.)} & SCIP-DEF       		& 1636.4 & 542.8  & 3.97& 198.5  \\
					 & ML-Split       	& 1630.1 & 202.7  & 3.62 & 202.8  \\
					 & PB-DFS      		& \textbf{1629.0} & \textbf{5.7}  & \textbf{3.46} & 5.7 \\ \cmidrule{1-6}

\multirow{3}{*}{DSP (Min.)} & SCIP-DEF       		& 315.5 & 390.8 & 3.0 & 114.2  \\
					 & ML-Split       	& \textbf{315.3} & 358.9  & \textbf{2.95} & 110.1  \\
					 & PB-DFS      		& 316.1 & 303.4  & 3.15 & 11.5  \\ \midrule

\multirow{3}{*}{MISP (Max.)} & SCIP-DEF       		& 1364.9 & 559.9  & 4.67 & 174.5 \\
					 & ML-Split       		& 1370.4 & 281.6 & 4.34 & 200.4 \\ 
					 & PB-DFS      	 & \textbf{1371.0} & \textbf{4.5} & \textbf{4.17} & 4.5  \\ \cmidrule{1-6}

\multirow{3}{*}{CAP (Max.)}  & SCIP-DEF     		& 3824.3 & 535.0  & 11.62 & 29.2 \\
					 & ML-Split       	& 3821.2 & 667.5   & 12.01 & 30.8\\
					 & PB-DFS      		& \textbf{3831.4} & \textbf{514.6}  & \textbf{11.35} & 5.5\\ 
\bottomrule

\end{tabular}}
\end{table}

We further demonstrate the effectiveness of PB-DFS by comparing the use of SCIP equipped with PB-DFS only, against both the full-fledged SCIP solver (SCIP-DEF) and a problem-reduction approach \cite{ding2019optimal} using SCIP-DEF as the solver (ML-Split). Figure \ref{fig:primalbound} presents the change of the primal bound during the solving process and the detailed solving statistics are shown in Table \ref{tab:heuristics2}. From Figure \ref{fig:primalbound}, we observe that PB-DFS finds (near-)optimal solutions at the very beginning and outperforms other methods on VCP and MISP. On DSP, early good solutions found by PB-DFS are still very useful such that it can help the solver without any primal heuristic outperform full-fledged solvers for a while. Further, PB-DFS is computationally cheap. Therefore, when incorporated into the SCIP solver, PB-DFS does not introduce a large computational overhead, and keeps more computational resources for the B\&B process. This explains why the quality of the incumbent solution quickly catches up with other approaches on CAP. The detailed solving statistics in Table \ref{tab:heuristics2}  are consistent with the above analysis, which confirms that the PB-DFS method is very competitive

%\comment[id=YS]{If so, what's the point of developing primal heuristics? Revise this sentence.} 

%We see that PB-DFS is a promising method for learning efficient primal heuristics, which allows the solver to devote more computational time on improving the dual bound under limited computational budgets.   

% \input{2-relatedwork}
\section{Conclusion}
\label{sec:conclusion}

In this work, we propose a primal heuristic based on machine learning, Probabilistic Branching with guided Depth-First Search (PB-DFS). PB-DFS is a B\&B configuration specialized for boosting the search for high-quality primal solutions, by leveraging a predicted solution to guide the search process of the B\&B method. Results show that PB-DFS can find better primal solutions, and find them much faster on several NP-hard covering problems as compared to other general heuristics in a state-of-the-art open-source MIP solver, SCIP. Further, we demonstrate that PB-DFS can make better use of high-quality predicted solutions as compared to recent solution prediction approaches.

We would like to note several promising directions beyond the scope of this work. Firstly, we demonstrate that PB-DFS can be deployed as a primal heuristic that runs only at the root node during a solver's B\&B process. More sophisticated implementations, e.g. triggering PB-DFS to run on different nodes with engineered timings, can lead to further performance improvement. Secondly, PB-DFS relies on high-quality predicted solutions. We observe the drop in the performances of existing ML models when extending it to general MIP problems. We expect that improving ML models in the context of solution prediction for Mixed Integer Programs could be a fruitful avenue for future research. 

% Secondly, both general heuristics in SCIP and PB-DFS do not work well for large-scale highly constrained problems, e.g., TSP with 200 cities. We would like to investigate these more and extend PB-DFS to this class of problem. 

%\todo{Conclusion is more like ``abstract'' but with an emphasis on our contributions and take-home message. I suggest to revise the conclusion in a way that people can understand the conclusion without referring to the main paper, because many people actually read conclusion first and then decide whether to read the whole paper. }
% \bibliographystyle{IEEEtran}
% \bibliography{lib}

\section*{Appendix A}
The features for decision variables are outlined as follows:

\begin{itemize}
    \item original, positive and negative objective coefficients; 
    \item number of non-zero, positive, and negative coefficients in constraints;
    \item variable LP solution of the original problem $\hat{x}$; $\hat{x} - \lfloor\hat{x}\rfloor$; $\lceil\hat{x}\rceil-\hat{x}$; a boolean indicator for whether $\hat{x}$ is fractional;
    \item variable's upward and downward pseudo costs; the ratio between these pseudocosts; sum and product of these pseudo costs; variable's reduced cost;
    \item global lower bound and upper bound;
    \item mean, standard deviation, minimum, and maximum degree for constraints in which the variable has a non-zero coefficient. The degree of a constraint is the number of non-zero coefficients of that constraint;
    \item the maximum and the minimum ratio between the left-hand-side and right-hand-side over constraints where the variable has a non-zero coefficient;
    \item statistics (sum, mean, standard deviation, maximum, minimum) for a variable's positive and negative constraint coefficients respectively;
    \item coefficient statistics of all variables in the constraints (sum, mean, standard deviation, maximum, minimum) with respect to three weighting schemes: unit weight, dual cost, the inverse of the sum of the coefficients in the constraint.
\end{itemize}

\section*{Appendix B}

The MIP formulations for tested problems are as follows:

\paragraph{Maximum Independent Set Problem (MISP)} In an undirected graph $\mathcal{G} (\mathcal{V}, \mathcal{E})$, a subset of nodes $\mathcal{S}$ is independent iff there is no edge between any pair of nodes in $\mathcal{S}$. A MISP is to find an independent set in $\mathcal{G}$ of maximum cardinality. The MIP formulation of the MISP is: $\max_{\mathbf{x}} \sum_{v \in \mathcal{V}}{x_v}$, subject to $x_u + x_v \le 1, \forall (u, v) \in \mathcal{E}$ and $x_v \in {0,1}, \forall v \in \mathcal{V}$.

\paragraph{Dominant Set Problem (DSP)} In an undirected graph $\mathcal{G} (\mathcal{V}, \mathcal{E})$, a subset of nodes $\mathcal{S} \subset \mathcal{V}$ dominates the complementary subset $ \mathcal{V} \setminus \mathcal{S}$ iff every node not in $\mathcal{S}$ is adjacent to at least one node in $\mathcal{S}$. The objective of a DSP is to find a dominant set in $\mathcal{G}$ of minimum cardinality. The MIP of the DSP is as follows: $\min_{\mathbf{x}} \sum_{v \in \mathcal{V}}{x_v}$, subject to $ x_v + \sum_{u \in N(v)} x_u \ge 1, \forall v \in \mathcal{V}$ and $x_v \in {0,1}, \forall v \in \mathcal{V}$. $N(v)$ denotes the set of neighborhood nodes of $v$.

\paragraph{Vertex Cover Problem (VCP)} In an undirected graph $\mathcal{G} (\mathcal{V}, \mathcal{E})$, a subset of nodes $S \subset V$ is a cover of $\mathcal{G}$ iff for every edge $e \in \mathcal{E}$, there is at least one endpoint in $S$. The objective of the VCP is to find a cover set in $\mathcal{G}$ of minimum cardinality. The MIP of the VCP is as follows: $\min_{\mathbf{x}} \sum_{v \in \mathcal{V}}{x_v}$, subject to $x_u + x_v \ge 1, \forall (u, v) \in \mathcal{E}$ and $x_v \in {0,1}, \forall v \in \mathcal{V}$.

\paragraph{Combinatorial Auction Problem (CAP)} A seller faces to selectively accept offers from bidders. Each offer indexed by $i$ contains a bid $p_i$ for a set of items (bundle) $\mathcal{C}_i$ by a particular bidder. The seller has limited amount of goods and aims to allocate the goods in a way that maximizes the total revenue. We use $\mathcal{I}$ and $\mathcal{J}$ to denote the index set of offers and the index set of items. Formally, the MIP formulation of the problem can be expressed as $\max_{\mathbf{x}} \sum_{i \in \mathcal{I}}{ p_i x_i}$, subject to $\sum_{k \in \{i \,|\, j \in \mathcal{C}_i, \forall i\}} x_k \le 1, \forall j \in \mathcal{J}$ and $x_i \in {0,1}, \forall i \in \mathcal{I}$.  

For MISP, DSP, and VCP, we sample random graphs using Erdős-Rényi generator. The affinity is set to 4. The training data consists of examples from solved small graph instances between 500 to 1001 nodes. We form the small-scale and medium-scale testing datasets with solved graph instances of 1000 nodes and those of 2000 nodes. We evaluate heuristics on large-scale graph instances of 3000 nodes. The CAP instances are generated using an arbitrary relationship procedure. The CAP instances in small-scale are sampled with items in the range [100, 150] and bids in the range [500, 750]. Instances in medium-scale and large-scale are generated with 150 items for 750 bids and 200 items for 1000 bids, respectively. The detailed parameter settings are given by following \cite{leyton2000towards}. 

% Table \ref{tab:mipstats} summarizes the detailed statistics from the MIP formulation of those tested problems. The $[\cdot, \cdot]$ denotes an interval from which a parameter is sampled. 

\section*{Appendix C}

\begin{table}[t!]
\centering
\caption{Effect of Score Functions}
\label{tab:scorefn}
\resizebox{0.95\columnwidth}{!}{
\begin{tabular}{@{}ccccccccc@{}}

    &   \multicolumn{2}{c}{Independent Set} & \multicolumn{2}{c}{Dominant Set} &   \multicolumn{2}{c}{Vertex Cover} & \multicolumn{2}{c}{Combinatorial Auction}\\
\toprule
 score function & objective & time & objective & time &  objective & time & objective & time \\
\midrule

 $ \mathbf{p} $ & 1371.0 & 2.8 & 318.7 & 9.9 & 1629.0 & 3.9   & 3226.6 & 2.3  \\
 $ \mathbf{1} - \mathbf{p} $  & 1371.1 & 3.7 & 322.0 & 11.3 & 1629.1 & 3.3 & 3317.0 & 5.2   \\
 $\max (\mathbf{p}, \mathbf{1}-\mathbf{p})$ & 1371.0 & 3.8 & 321.9 & 11.6 &  1628.9 & 3.9  & 3316.0 & 5.2 \\
 
\bottomrule
\end{tabular}}
\end{table}

%  In the main paper, we have demonstrated the score function $z_i \gets \max(p_i, 1-p_i)$ without discussing other possibilities. Here, we consider two alternative score functions, 
 
 We consider two alternative score functions, $z_i \gets p_i$ and $z_i \gets 1-p_i$, subject to $i \in \mathcal{C}$. When using $z_i \gets p_i$ as the score function to select the decision variable with the maximum score to branch, our guided DFS always prefers the node that is the result of fixing the decision variable to $1$. The behavior of the PB-DFS due to this score function can be interpreted as incrementally adding variables that more likely belong to an optimal solution until a feasible solution is obtained. When using the other score function $z_i \gets 1 - p_i$, guided DFS always prefers the node that is the result of fixing the decision variable to $0$, and the resulting behavior of the PB-DFS can be interpreted as continuously removing variables that less likely belong to an optimal solution until a feasible solution is obtained. In table \ref{tab:scorefn}, we observe that the score functions do not significantly affect the first-found solution. 

\end{document}